\documentclass[]{unireason}
\usepackage{fix-cm}
\usepackage{helvet}           

\usepackage{nicefrac}         
\usepackage{siunitx}         

\usepackage{longtable}
\usepackage{booktabs}

\usepackage{tabularx}         
\usepackage{array}           
\usepackage{makecell}         
\usepackage{threeparttable}

\usepackage{wrapfig}          
\usepackage{float}            
\usepackage[export]{adjustbox}

\usepackage{tikz}             
\usepackage{pgfplots}         
\usepackage{pgf-pie}

\usepackage{listings}        
\usepackage{ragged2e}        
\usepackage{comment}

\usepackage{pifont}           
\usepackage{fontawesome}

\usepackage{enumitem}        
\usepackage{titletoc}         
\usepackage{minitoc}          
\usepackage[toc,page,header]{appendix}

\usepackage{url}              
\usepackage[hang,flushmargin]{footmisc}  
\pgfplotsset{compat=1.18}

\usepackage{xspace}
\newcommand{\methodname}{UniReason\xspace}
\newcommand{\modelname}{UniReason\xspace}

\definecolor{lightblue}{RGB}{200, 230, 255}  
\definecolor{headerblue}{RGB}{150, 200, 255}

\usepackage{multirow}
\usepackage{xcolor,colortbl}
\definecolor{lavendergray}{rgb}{0.77, 0.76, 0.82}
\definecolor{lightgray}{rgb}{0.83, 0.83, 0.83}

\usepackage{array}
\newcolumntype{H}{>{\setbox0=\hbox\bgroup}c<{\egroup}@{}}
\newcommand{\tablestyle}[2]{\setlength{\tabcolsep}{#1}\renewcommand{\arraystretch}{#2}\centering\small}

\usepackage{amssymb}%
\usepackage{pifont}%

\usepackage{wrapfig}

\definecolor{myblue}{rgb}{0.11764705882352941, 0.5647058823529412, 1.0}
\definecolor{Gray}{gray}{0.9}
\definecolor{darkgreen}{rgb}{0.545, 0.749, 0.608}
\newcommand{\gain}[1]{{\color{darkgreen}\scriptsize\textbf{#1}}}

\usepackage{tablefootnote}

\usepackage{soul}
\sethlcolor{Gray}

\newcounter{examplebox}



\title{%
  \begin{minipage}[c]{0.08\textwidth}
    \includegraphics[height=2.8em]{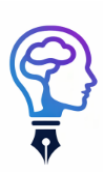} 
  \end{minipage}%
  \hspace{0.1em}%
  \begin{minipage}[c]{0.90\textwidth}
    \centering
    \textbf{\modelname} 1.0: A Unified Reasoning Framework for \\ World Knowledge Aligned \\ Image Generation and Editing
  \end{minipage}
}

\author{
    Dianyi Wang\textsuperscript{1,2*}, 
    Chaofan Ma\textsuperscript{3*},   
    Feng Han\textsuperscript{1,2},
    Size Wu\textsuperscript{4}, \\
    Wei Song\textsuperscript{2,5},
    Yibin Wang\textsuperscript{1,2}, 
    Zhixiong Zhang\textsuperscript{2,3},
    Tianhang Wang\textsuperscript{2,5}, \\
    Siyuan Wang\textsuperscript{6$\dagger$},
    Zhongyu Wei\textsuperscript{1,2$\dagger$},
    Jiaqi Wang\textsuperscript{2$\dagger$}
}

\affiliation[1]{\mbox{Fudan University}}
\affiliation[2]{\mbox{Shanghai Innovation Institute}}
\affiliation[3]{\mbox{Shanghai Jiao Tong University}}
\affiliation[4]{\mbox{Nanyang Technological University}}
\affiliation[5]{\mbox{Zhejiang University}}
\affiliation[6]{\mbox{University of Southern California}}

\contribution[\textbf{*}]{Equal Contribution}
\contribution[\dagger]{Corresponding Authors}

\abstract{
    Unified multimodal models often struggle with complex synthesis tasks that demand deep reasoning, and typically treat text-to-image generation and image editing as isolated capabilities rather than interconnected reasoning steps. To address this, we propose \textbf{\modelname}, a unified framework that harmonizes these two tasks through two complementary reasoning paradigms. We incorporate \textit{world knowledge-enhanced textual reasoning} into generation to infer implicit knowledge, and leverage editing capabilities for \textit{fine-grained editing-like visual refinement} to further correct visual errors via self-reflection. This approach unifies generation and editing within a shared architecture, mirroring the human cognitive process of planning followed by refinement. We support this framework by systematically constructing a large-scale reasoning-centric dataset ($\sim$300k samples) covering five major knowledge domains (\textit{e.g.}, cultural commonsense, physics, etc.) for textual reasoning, alongside an agent-generated corpus for visual refinement. Extensive experiments demonstrate that \textbf{\modelname} achieves advanced performance on reasoning-intensive benchmarks such as WISE, KrisBench and UniREditBench, while maintaining superior general synthesis capabilities.
}

\checkdata[GitHub]{\url{https://github.com/AlenjandroWang/UniReason}}
\checkdata[HuggingFace]{\url{https://huggingface.co/Alex11556666/UniReason}}

\begin{document}
\maketitle


\vspace{-1.5em}

\section{Introduction}
\label{sec:intro}

\begin{figure}[t]
    \centering
    \includegraphics[width=0.92\linewidth]{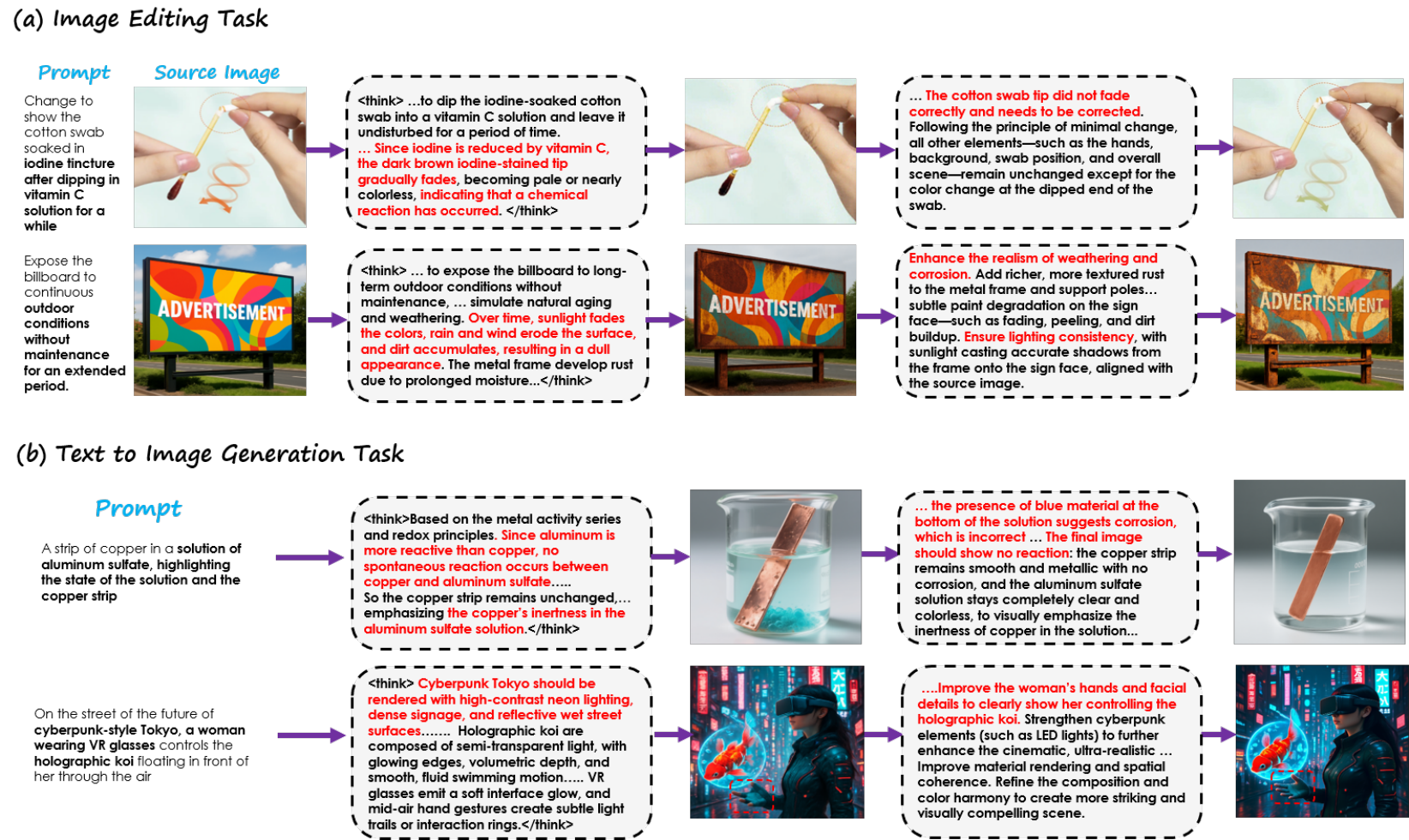} 
    \caption{Illustrative cases of \textbf{\methodname} on \textit{image editing} and \textit{T2I generation} tasks.
Given an instruction, the model first performs world knowledge-enhanced textual reasoning to generate grounded, fine-grained guidance for image synthesis. It then applies fine-grained editing-like visual refinement, correcting errors introduced during the initial generation and improving the synthesis quality.}
    \label{Fig:teaser}
\end{figure}

Unified multimodal models have emerged as a promising paradigm for jointly handling visual understanding and generation tasks~\cite{xie2025showosingletransformerunify, chen2025janusprounifiedmultimodalunderstanding, wu2025harmonizing, wu2024vila, pan2025transfer, deng2025emergingpropertiesunifiedmultimodal}.
By integrating perception and synthesis within a shared architecture, these models enable seamless interplay between comprehending visual content and producing new images conditioned on multimodal inputs. 
Among various capabilities, \textit{text-to-image (T2I) generation} and \textit{image editing} stand out as particularly challenging yet impactful applications. 
However, current unified models still struggle with complex scenarios, where tasks demand not only precise instruction following, but also world knowledge that extends beyond surface-level pixels, \textit{e.g.}, commonsense, physical laws, and spatial-temporal logic.
Such challenges fundamentally demand \textit{reasoning} capabilities to bridge the gap between abstract user intent and faithful visual output.

To enhance reasoning capabilities, a prominent line of work focuses on \textit{prompt enhancement} or \textit{reprompting} strategies~\cite{jiang2025t2ir1reinforcingimagegeneration,fang2025gotunleashingreasoningcapability,xiao2025mindomniunleashingreasoninggeneration}. 
These methods employ chain-of-thought (CoT) reasoning to expand abstract user prompts into explicit semantic and spatial guidance before generation.
While effective in improving instruction alignment, these ``reason-then-generate'' approaches are inherently limited as reasoning occurs only \textit{before} generation without access to visual feedback, preventing reflection on and correction of output errors. More recently, \textit{interleaved reasoning} mechanisms ~\cite{huang2025interleavingreasoningbettertexttoimage, qin2025unicotunifiedchainofthoughtreasoning} alternate between textual reasoning and visual generation. 
By first generating an initial image, then performing textual reflection based on visual feedback, and finally refining the output, these approaches enable post-generation correction that was previously infeasible.

Despite this progress, existing methods still exhibit two key limitations. 
(1) Reasoning in these methods largely remains at the level of semantic reorganization, decomposing instructions into finer-grained descriptions or spatial layouts~\cite{xiao2025mindomniunleashingreasoninggeneration,jiang2025t2ir1reinforcingimagegeneration,qin2025unicotunifiedchainofthoughtreasoning,ghosh2023genevalobjectfocusedframeworkevaluating}. 
This addresses only the explicit component of user intent, whereas faithful synthesis in practice demands \textit{world knowledge} that is implicitly assumed rather than explicitly stated. 
Such knowledge must be \textit{inferred}, not merely \textit{parsed} from instructions. 
This creates a fundamental \textit{knowledge gap} that surface-level decomposition cannot bridge.
(2) Existing methods typically address text-to-image generation and image editing as separate tasks~\cite{huang2025interleavingreasoningbettertexttoimage}, leaving their inherent synergies within a unified interleaved framework untapped. 
We argue that these two tasks share substantial reasoning overlap and can mutually reinforce each other. Specifically, post-generation critique and refinement in interleaved reasoning is structurally analogous to editing. Isolating them therefore forgoes such synergy and leads to redundant learning.

To address these challenges, we propose \textbf{\modelname}, a unified reasoning framework that harmonizes text-to-image generation and image editing within a shared architecture, as illustrated in Fig.~\ref{Fig:teaser}. Our framework supports two complementary reasoning paradigms.
\textit{\textbf{(1) World Knowledge-Enhanced Textual Reasoning}} aims to bridge the knowledge gap prior to synthesis. Given an underspecified instruction, the model performs textual reasoning to infer implicit world knowledge and produces grounded guidance that specifies fine-grained details for the subsequent image synthesis. To support this, we construct training data across five knowledge categories: cultural commonsense, natural science, spatial, temporal, and logical reasoning. We use large language models to generate reasoning traces and apply multi-dimensional filtering to ensure high-quality supervision.
\textit{\textbf{(2) Fine-grained Editing-like Visual Refinement}} aims to improve synthesis quality after initial generation. Given the initial image and prior reasoning, the model performs self-reflection to identify discrepancies or missing details, then applies targeted corrections to produce a refined image. Observing that this process is structurally analogous to image editing, we jointly learn T2I generation and editing for mutual benefit. We design an agent pipeline that iterates through generation, verification, refinement, and comparison to construct high-quality training data.
These two paradigms can be applied independently or jointly, offering flexibility across diverse synthesis scenarios. 

We adopt a two-stage training strategy: the first stage strengthens foundational generation capability, and the second stage enables interleaved reasoning by jointly training the understanding and generation branches.
Through this unified framework, we achieve comprehensive world knowledge-grounded reasoning capabilities for both T2I generation and image editing, with advanced performance on multiple benchmarks.

Our main contributions are summarized as follows:
\begin{itemize}
    \item We propose \textbf{\modelname}, a unified \textit{reasoning} framework for both T2I generation and image editing. Our key insight is that refinement and editing share the same reasoning pattern, enabling bidirectional capability transfer.
    \item We introduce two complementary reasoning paradigms: \textit{World Knowledge-Enhanced Textual Reasoning} bridges the knowledge gap before synthesis, while \textit{Fine-grained Editing-like Visual Refinement} enables iterative improvement after generation.
    \item We systematically construct training data for both paradigms, including world knowledge-aligned data across five categories and an agent pipeline for refinement supervision, combined with a two-stage training strategy.
    \item Extensive experiments demonstrate advanced performance on multiple benchmarks, including GenEval, WISE for T2I generation, and UniREditBench, KrisBench for image editing.
\end{itemize}

\section{Related Work}

\paragraph{Image Generation and Editing}
Image generation (T2I) and editing are two related tasks, depending on whether the conditional signals are textual descriptions or reference images.
Recently, Diffusion Transformers~\cite{vaswani2017attention, peebles2023scalable} (DiTs) have served as the backbone of state-of-the-art generation frameworks, with flow-matching~\cite{esser2024scaling,ma2024sit} adopted as the prevailing training scheme.
Together with data and model scaling, these advances have enabled photo-realistic synthesis and substantially improved instruction following in T2I generation~\cite{esser2024scaling, wu2025qwenimagetechnicalreport, blackforest2024flux}. Building upon these powerful generators, recent image editing systems~\cite{labs2025flux, wu2025qwenimagetechnicalreport} achieve precise content manipulation while preserving overall visual consistency. 
However, despite their generative prowess, these specialized models lack the intrinsic capacity for world comprehension and self-reflection, motivating the integration of reasoning and generation within a coherent unified framework.

\paragraph{Unified Multimodal Models}
Unified multimodal models~\cite{wu2025harmonizing, wu2024vila, pan2025transfer, chen2025blip3ofamilyfullyopen, deng2025emergingpropertiesunifiedmultimodal, chen2025janusprounifiedmultimodalunderstanding} aim to jointly support image understanding and generation within a single framework. Broadly, existing approaches can be grouped into two paradigms.
A first, more modular paradigm aligns pretrained LMMs and DiTs via LLM hidden states~\cite{wu2025harmonizing, lin2025uniworldv1highresolutionsemanticencoders, wu2025omnigen2explorationadvancedmultimodal} or learnable queries~\cite{chen2025blip3ofamilyfullyopen, pan2025transfer, wu2025openuni,wang2026deepgen10alightweightunified}. 
Another line of work~\cite{xie2025showosingletransformerunify, chen2025janusprounifiedmultimodalunderstanding, deng2025emergingpropertiesunifiedmultimodal,wang2025autoregressive} adopts a shared LLM architecture for perception and synthesis, encouraging a tight coupling between the two tasks. In our study, we focus on the second paradigm, since a shared backbone naturally supports interleaved reasoning between language and image generation in a unified inference process.

\paragraph{Reasoning in Unified Multimodal Models} The structural convergence of understanding and generation within unified models unlocks the potential for grounding high-fidelity image synthesis in complex multimodal reasoning. Initial efforts primarily involve the adaptation of textual Chain-of-Thought (CoT) to image generation~\cite{jiang2025t2ir1reinforcingimagegeneration,fang2025gotunleashingreasoningcapability,xiao2025mindomniunleashingreasoninggeneration}, following a ``reason-then-generate'' paradigm that expands user instructions into detailed descriptions prior to synthesis. More recently, interleaved reasoning mechanisms~\cite{huang2025interleavingreasoningbettertexttoimage, qin2025unicotunifiedchainofthoughtreasoning} extend the process into iterative "reason-generate-reflect" cycles to incorporate visual feedback. Despite these advancements, existing methods are often confined to prompt reorganization and rigidly separate generation and editing tasks. In this work, we address these limitations by inferring implicit world knowledge rather than merely parsing instructions. Furthermore, we exploit the inherent synergies between T2I generation and image editing within a unified reasoning framework.

\section{Preliminary}
\paragraph{Architecture} We build upon Bagel~\cite{deng2025emergingpropertiesunifiedmultimodal} to develop a unified and interleaved reasoning framework for both T2I generation and image editing. Bagel adopts a Mixture-of-Transformers (MoT) architecture with a ViT encoder~\cite{tschannen2025siglip2multilingualvisionlanguage} to process multimodal inputs and enables unified image understanding and generation within a single foundation model.

Specifically, multimodal understanding is formulated as generating context-aware textual outputs via standard next-token prediction through a language modeling head. This process is conditioned on multimodal context inputs and handled by the understanding expert. Formally, the training objective minimizes the negative log-likelihood:
\begin{equation}
\centering
\mathcal{L}_{\text{text}} = - \sum_{t=1}^{T} \log p_{\theta}\!\left(x_t \mid x_{<t}, C\right),
\end{equation}
where $x_t$ denotes the target text token, $x_{<t}$ is the preceding tokens and $C$ is the multimodal context.

Multimodal generation focuses on producing high-quality and semantically aligned images via a rectified flow process~\cite{liu2022flowstraightfastlearning} in a VAE's latent space~\cite{blackforest2024flux}, conditioned on multimodal inputs and handled by the generation expert. The training objective is to minimize the the latent flow-matching loss:
\begin{equation}
\centering
\mathcal{L}_{\text{image}}
= \mathbb{E}_{t \sim \mathcal{U}(0,1)}
\left\|
u_{\theta}\!\left(z_t, t \,;\,C\right)
- u^{\star}\!\left(z_t, t\right)
\right\|_2^2,
\end{equation}
where $u^{\star}$ denotes the target velocity, $u_{\theta}$ is the learned time-conditioned velocity field in the latent space.

\paragraph{Reasoning Paradigms} Bagel's unified architecture supports interleaving textual reasoning and visual synthesis in both T2I generation and image editing tasks. Specifically, T2I generation takes a textual instruction as input and outputs a sequence of intermediate reasoning tokens together with a synthesized image. For image editing, an existing image and a textual instruction are taken as input and the model outputs a reasoning text and the edited image.
In this work, we formulate interleaved reasoning as an iterative process: 
$
(I^{k+1}, T^{k+1}) = \mathcal{F}\big(I^{\leq k}, T^{\leq k}, C\big)
$
where $I^{k}$ and $T^{k}$ denote the image and reasoning text at iteration $k$, $C$ denotes the multimodal context, and $\mathcal{F}$ is the unified model ($k=1$ in our implementation). 
Under this formulation, each refinement step can be interpreted as an image editing operation conditioned on the reasoning trace. Therefore, we propose to jointly learn T2I generation and image editing within a unified interleaved reasoning framework, allowing the refinement process to benefit from editing learning and, conversely, enhance interleaved reasoning for both T2I generation and editing.

\section{Method}
\label{Framework}
\begin{figure*}[t]
    \centering
    \includegraphics[width=0.92\linewidth]{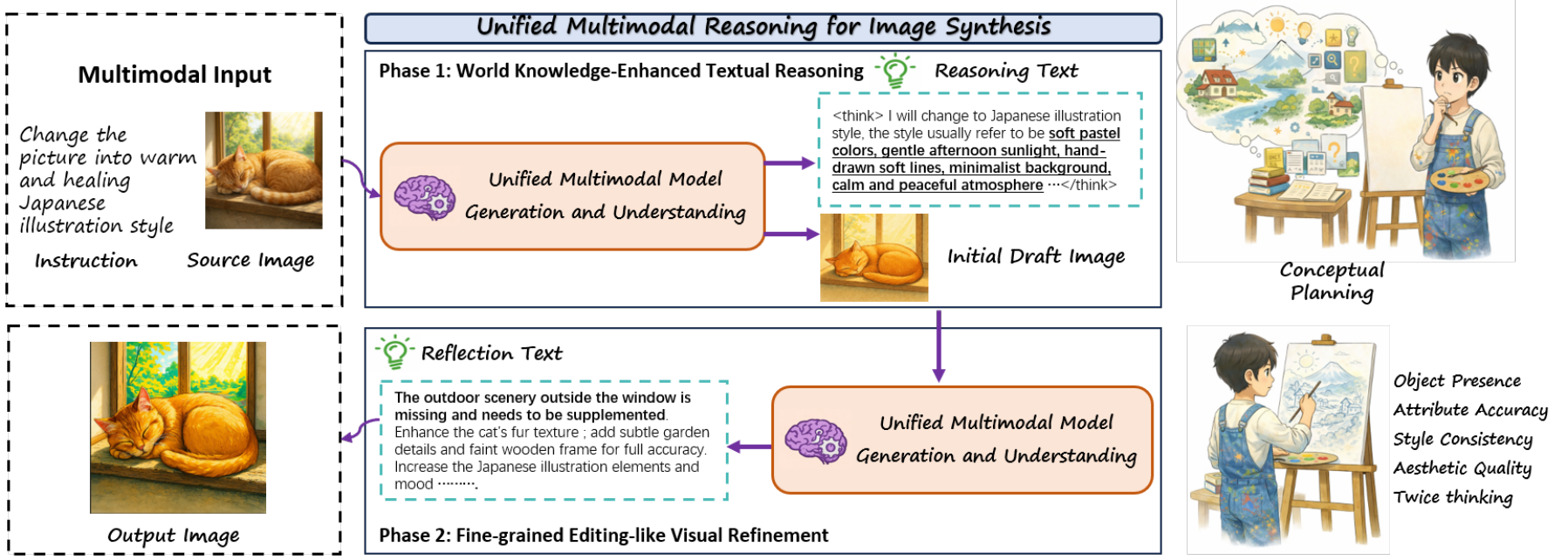} 
    \caption{Overview of \textbf{UniReason} framework for two complementary
reasoning paradigms in image synthesis.}
    \label{Fig:method}
\end{figure*}

\begin{figure*}[th]
    \centering
    \includegraphics[width=0.92\linewidth]{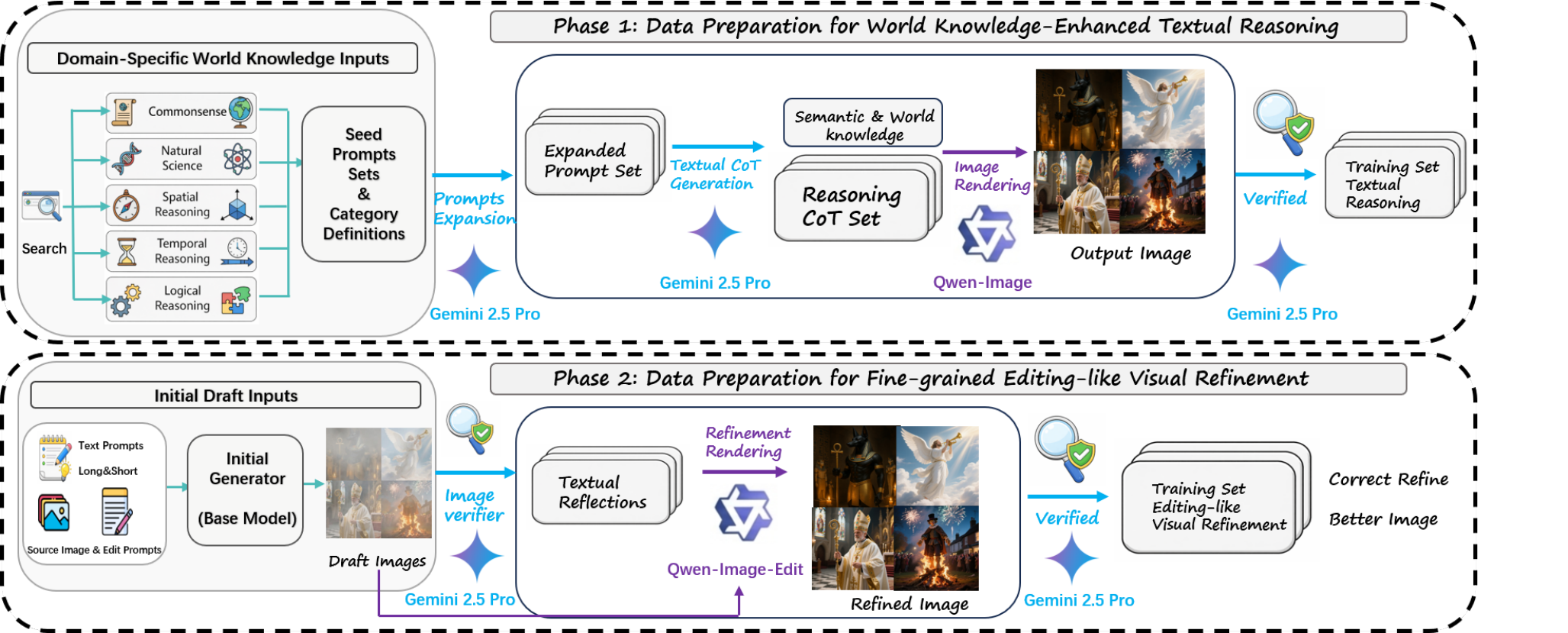} 
    \caption{Overview of our data preparation framework.}
    \label{Fig:data}
\end{figure*}

In this section, we present \textbf{UniReason}, a unified multimodal reasoning framework for both T2I generation and image editing, 
as illustrated in Fig.~\ref{Fig:method}. In practice, the framework operates in two phases, (1) \textit{World Knowledge-Enhanced Textual Reasoning} for initial synthesis; (2) \textit{Fine-grained Editing-like Visual Refinement} for iterative improvement. We introduce each phase along with its corresponding data creation pipeline in Sec.~\ref{world_knowledge_reasoning} and~\ref{visual_refinement}, respectively shown in Fig.~\ref{Fig:data}, followed by the training strategy in Sec.~\ref{sec:training recipe}.

\subsection{World Knowledge-Enhanced Textual Reasoning} 
\label{world_knowledge_reasoning}
Different from prior work~\cite{fang2025gotunleashingreasoningcapability,xiao2025mindomniunleashingreasoninggeneration} that primarily focuses on re-organizing user instructions into more detailed visual descriptions, our core objective is to enable the unified multimodal model to not only expand raw user prompts but also understand the underlying implicit world knowledge. Specifically, \textbf{UniReason} utilizes textual reasoning to infer the world knowledge required to complete the visual synthesis, including commonsense, cultural context, time-spatial and natural science principles. This process provides explicit and structured guidance to ensure the initial generation is both instruction-aligned and knowledge-consistent, mirroring the conceptual planning that humans perform when outlining ideas for a drawing.

\paragraph{Data Preparation} To enable world knowledge-enhanced textual reasoning for the initial synthesis, we construct challenging input instructions for both T2I generation and image editing tasks that require complex world knowledge reasoning beyond complementing pixel-level details, along with their associated reasoning processes. Specifically, we cover five major categories of world knowledge and adopt post-generation filtering to ensure high-quality supervision.
\begin{itemize}
    \item \textbf{Cultural Commonsense} instructions require using shared cultural knowledge, such as historical events, iconic figures, social customs, and idiomatic expressions, to resolve unnamed or underspecified entities into explicit, contextually meaningful visual content, ensuring generated images aligned with real-world cultural understanding. 
    \item \textbf{Natural Science} instructions requires incorporating principles from physics, biology, medicine, or chemistry to ensure that generated images remain consistent with scientific laws, and reflect plausible real-world observations. 
    \item \textbf{Spatial} reasoning focuses on understanding correct spatial relationships among entities, including relative position, orientation, viewpoint, and camera transformations. Such instructions requires deriving precise spatial configurations from abstract descriptions to generate visuals consistent with real-world geometric logic.
    \item \textbf{Temporal} reasoning models time-dependent relationships, such as event sequences, state transitions, and causal ordering. This type of instructions require inferring the temporal progression of events and ensuring that visual outputs reflect coherent and plausible temporal dynamics aligned with natural chronological flow.
    \item \textbf{Logical} reasoning emphasizes causal coherence and logical consistency during image generation, such as in maze-solving or constraint satisfaction problems, by adhering to explicit or implicit logical structures. These instructions require applying deductive principles to translate abstract logical constraints into visually valid solutions.
\end{itemize}

For T2I generation in each category, we manually construct seed prompts based on Wikipedia, together with explicit category definitions, and use Gemini-2.5 Pro~\cite{google_gemini25_pro_2025} to expand them into a larger prompt set. And Gemini-2.5 Pro is also employed to generate textual CoT reasoning for each prompt. All prompts with their corresponding CoTs are subsequently fed into Qwen-Image~\cite{wu2025qwenimagetechnicalreport} for image rendering to form paired training samples. For image editing, we utilize data triples (original image, editing instruction, desired outcome) from UniREdit-Data-100K~\cite{han2025unireditbenchunifiedreasoningbasedimage} that covers diverse knowledge dimensions, and expand them with textual reasoning traces generated by Gemini-2.5 Pro. Moreover, to ensure the training samples are generated without hallucinations, Gemini-2.5 Pro serves as a comprehensive evaluator to assess the generated images across three dimensions: instruction alignment, visual fidelity, and reasoning correctness. Only verified samples are retained to construct a high-quality training set for training visual synthesis with textual reasoning.

\subsection{Fine-grained Editing-like Visual Refinement}
\label{visual_refinement}
After the initial visual generation or editing, the draft already captures essential elements and semantically aligns with the input instruction and world knowledge, but inevitably contains imperfections that require fine-grained refinement. We therefore continue refining the results from the knowledge-enhanced initial synthesis. Specifically, the model reassesses the initial synthesized image considering prior textual reasoning, reflectively identifies and verbalizes inconsistencies and missing details. It then optionally incorporates a second round of textual reasoning, which accordingly refines semantic attributes, aesthetic details, stylistic coherence and instruction consistency to produce a polished image.  
This refinement process guided by textual reflection is structurally analogous to image editing, motivating us to create a synergistic loop for mutual improvement between T2I generation and image editing, by alternating knowledge-enhanced textual reasoning and editing-like visual refinement.

\paragraph{Data Preparation} We design an agent pipeline to construct high-quality supervision data for training interleaved reasoning across both T2I generation and image editing tasks. The pipeline consists of (i) an initial generator (the base model) that produces a draft image with its textual reasoning from the input; (ii) a verifier (Gemini-2.5 Pro) that diagnoses caption–image mismatches and outputs structured, actionable edit directives across five dimensions: object presence, attribute accuracy, style consistency, realism, and aesthetic quality; (iii) a refinement teacher (Qwen-Image-Edit~\cite{wu2025qwenimagetechnicalreport}) that applies the feedback and textual reasoning via instruction-guided image editing to obtain an improved image; and (iv) a final judge (Gemini-2.5 Pro) that performs comparative evaluation between the initial and refined images, retaining refined images only if they exhibit measurable improvements over the initial generation and faithfully reflects the verifier's suggestion.

Specifically, we sample long-form captions from ShareGPT-4o-Image dataset~\cite{chen2025sharegpt4oimagealigningmultimodalmodels} and short-form captions from midjourney prompts\footnote{https://huggingface.co/datasets/vivym/midjourney-prompts} for T2I generation, and image-instruction pairs from UniREdit-Data-100K~\cite{han2025unireditbenchunifiedreasoningbasedimage} for image editing. These inputs are fed to the initial generator for reasoning-augmented initial synthesis. The caption–image pairs then undergo the full verification, refinement and comparison cycle, resulting in a corpus of high-quality training data for image synthesis with multimodal interleaved reasoning.

\subsection{Two-stage Training Strategy}
\label{sec:training recipe}
We adopt a simple yet effective two-stage supervised fine-tuning (SFT) strategy to first strengthen the foundational generation capability of the unified multimodal model, then 
train interleaved knowledge-enhanced reasoning and refinement capabilities across diverse image synthesis queries. 

\paragraph{Stage 1: Foundational Generation Strengthening} In the first stage, we freeze the multimodal understanding branch of the base model and train only the generation branch. This stage focuses exclusively on image synthesis using existing T2I generation and image editing datasets without textual reasoning, aiming to enhance the instruction-following ability and foundational image synthesis capability.

\paragraph{Stage 2: Interleaved Reasoning Tuning} In the second stage, we unfreeze all model parameters and jointly train the understanding and generation branches using the curated interleaved reasoning data, including single-turn knowledge-enhanced reasoning samples and iterative visual refinement samples. This enables the model to perform world knowledge-enhanced reasoning and iteratively reflect and refine visual content.
Specifically, for single-turn reasoning data, we supervise both the textual reasoning traces and the image synthesis outputs. For visual refinement data, we supervise textual reflections and refined images while leaving the initial reasoning text and visual draft unsupervised.
The overall objective is formulated as
\begin{equation}
\centering
\mathcal{L}
= \lambda_{\text{text}} \, \mathcal{L}_{\text{text}}
+ \lambda_{\text{img}} \, \mathcal{L}_{\text{img}},
\end{equation}
where $\mathcal{L}_{\text{text}}$ denotes the text loss for supervising the reasoning tokens, and $\mathcal{L}_{\text{img}}$ denotes the image loss for supervising the synthesized images. $\lambda_{\text{text}}$ and $\lambda_{\text{img}}$ are scalar loss weights that balance the contributions of the text and image objectives, respectively.

\begin{table*}[th!]
\centering
\caption{Evaluation of world knowledge-intensive text-to-image generation on the WISE~\cite{niu2025wiseworldknowledgeinformedsemantic} benchmark. "*" denotes generation with textual reasoning only, "†" denotes generation with both reasoning and refinement. The first block reports the performance of closed-source models. Bold entries represent the best performance among open-source models.}
\setlength\tabcolsep{12pt}
\resizebox{0.8\textwidth}{!}{
\begin{tabular}{cccccccc}
\toprule
\multicolumn{1}{c|}{Model}                        & Cultural                    & Time                        & Space                       & Biology                     & Physics                     & \multicolumn{1}{c|}{Chemistry}                   & Overall                   \\ \midrule
\multicolumn{1}{c|}{{\color[HTML]{1F2329} GPT-4o}} & {\color[HTML]{1F2329} 0.81} & {\color[HTML]{1F2329} 0.71} & {\color[HTML]{1F2329} 0.89} & {\color[HTML]{1F2329} 0.83} & {\color[HTML]{1F2329} 0.79} & \multicolumn{1}{c|}{{\color[HTML]{1F2329} 0.74}} & {\color[HTML]{1F2329} 0.80} \\ 
\multicolumn{1}{c|}{{\color[HTML]{1F2329} Seedream 4.0}} & {\color[HTML]{1F2329} 0.78} & {\color[HTML]{1F2329} 0.73} & {\color[HTML]{1F2329} 0.85} & {\color[HTML]{1F2329} 0.79} & {\color[HTML]{1F2329}  0.84} & \multicolumn{1}{c|}{{\color[HTML]{1F2329} 0.67}} & {\color[HTML]{1F2329} 0.78} \\\midrule
\rowcolor[HTML]{F6D6D3}\multicolumn{8}{c}{Unified Understanding and Generation w/o Reasoning.}                                                                                                                                                                                                                                \\ \midrule
\multicolumn{1}{c|}{Harmon}                        & 0.38                        & 0.48                        & 0.52                        & 0.37                        & 0.44                        & \multicolumn{1}{c|}{0.29}                        & 0.41                       \\
\multicolumn{1}{c|}{Show-o}                        & 0.28                        & 0.40                         & 0.48                        & 0.30                         & 0.46                        & \multicolumn{1}{c|}{0.30}                         & 0.35                       \\
\multicolumn{1}{c|}{Janus Pro}                     & 0.30                         & 0.37                        & 0.49                        & 0.36                        & 0.42                        & \multicolumn{1}{c|}{0.26}                        & 0.35                       \\
\multicolumn{1}{c|}{MetaQuery-XL}                  & 0.56                        & 0.55                        & 0.62                        & 0.49                        & 0.63                        & \multicolumn{1}{c|}{0.41}                        & 0.55                       \\
\multicolumn{1}{c|}{BLIP3-o}                       & –                           & –                           & –                           & –                           & –                           & \multicolumn{1}{c|}{–}                           & 0.62                       \\
\multicolumn{1}{c|}{UniWorld-V1}                   & 0.53                        & 0.55                        & 0.73                        & 0.45                        & 0.59                        & \multicolumn{1}{c|}{0.41}                        & 0.55                       \\
\multicolumn{1}{c|}{OmniGen2}                      & 0.42                        & 0.52                        & 0.64                        & 0.43                        & 0.50                         & \multicolumn{1}{c|}{0.34}                        & 0.47                       \\ 
\multicolumn{1}{c|}{Hunyuan-Image 3.0}                      & 0.58                        & 0.57                        &  0.70                        & 0.56                        &  0.63                         & \multicolumn{1}{c|}{ 0.31}                        & 0.57                       \\
\multicolumn{1}{c|}{Qwen-Image}                      & 0.62                        & 0.63                        & 0.77                        & 0.57                        &0.75                         & \multicolumn{1}{c|}{0.40}                        & 0.62                       \\\midrule
\rowcolor[HTML]{DCEBFA}\multicolumn{8}{c}{Unified Understanding and Generation w/ Reasoning.}                                                                                                                                                                                                       \\ \midrule
\multicolumn{1}{c|}{T2I-R1\textsuperscript{*}}                        & 0.56                      & 0.55                     & 0.63                      & 0.54                      & 0.55                      & \multicolumn{1}{c|}{0.30}                       & 0.54                   \\
\multicolumn{1}{c|}{MindOmni\textsuperscript{*}}                      & 0.75                    & 0.70                   & 0.76                   & 0.76                   & 0.72                   & \multicolumn{1}{c|}{0.52}                   & 0.71                   \\
\multicolumn{1}{c|}{IRG\textsuperscript{†}}                           & 0.78                      & \textbf{0.72}                      & 0.76                      & \textbf{0.81}                      &0.82                      & \multicolumn{1}{c|}{0.78}                      &0.77                     \\
\multicolumn{1}{c|}{BAGEL\textsuperscript{*}}                         & 0.76                   &0.69                   &0.75                   &0.65                   & 0.75                    & \multicolumn{1}{c|}{0.58}                   & 0.70                  \\
\multicolumn{1}{c|}{UniCoT\textsuperscript{†}}                        &0.76                   &0.70                   &0.76                   &0.73                   &0.81                   & \multicolumn{1}{c|}{0.73}                   &0.75                  \\
\multicolumn{1}{c|}{Ours\textsuperscript{†}}                          &\textbf{0.80}                   &0.68                   &\textbf{0.79}                    &0.77                   &\textbf{0.83}                    & \multicolumn{1}{c|}{\textbf{0.81}}                   &\textbf{0.78}                  \\ \midrule
\end{tabular}}
\label{table:wise}
\end{table*}

\begin{table*}[th!]
\centering
\caption{Evaluation of knowledge-intensive image editing on KrisBench~\cite{ye2025imgeditunifiedimageediting} and UniREditBench~\cite{liu2025step1xeditpracticalframeworkgeneral} benchmarks. "*" denotes textual reasoning only for editing, "†" denotes interleaved reasoning with both reasoning and refinement. Bold entries represent the best performance among open-source models.}
\setlength\tabcolsep{10pt}
\resizebox{0.85\textwidth}{!}{
\begin{tabular}{cccccccc}
\toprule
\multicolumn{1}{c|}{}                         & \multicolumn{4}{c|}{KrisBench}                                                                                                  & \multicolumn{3}{c}{UniREditBench}                                             \\ \cline{2-8} 
\multicolumn{1}{c|}{\multirow{-2}{*}{Model}} & Factual & Conceptual & \multicolumn{1}{c|}{Extract Procedural} & \multicolumn{1}{c|}{Overall}     & Real World & \multicolumn{1}{c|}{Game World} & Overall      \\ \cline{1-8} 
\multicolumn{1}{c|}{GPT-4o}                   & 79.80              & 81.37                & \multicolumn{1}{c|}{78.32}                        & \multicolumn{1}{c|}{80.09}       & 81.01               & \multicolumn{1}{c|}{62.07}               & 73.39        \\
\multicolumn{1}{c|}{Gemini 2.0}               & 65.26             & 59.65                & \multicolumn{1}{c|}{62.90}                         & \multicolumn{1}{c|}{62.41}       & –                   & \multicolumn{1}{c|}{–}                   & –            \\ 
\multicolumn{1}{c|}{Seedream 4.0}                   & –              & –                & \multicolumn{1}{c|}{–}                        & \multicolumn{1}{c|}{–}       & 66.22               & \multicolumn{1}{c|}{45.38}               & 55.77        \\\midrule
\rowcolor[HTML]{F6D6D3}\multicolumn{8}{c}{Unified Understanding and Generation w/o Reasoning.}                                                                                                                                                                                                       \\ \midrule
\multicolumn{1}{c|}{OmniGen2}                 & 57.36             & 44.20                 & \multicolumn{1}{c|}{47.79}                        & \multicolumn{1}{c|}{49.71}       & 53.69               & \multicolumn{1}{c|}{33.14}               & 43.41        \\
\multicolumn{1}{c|}{Uniworld V1}              & 47.71             & 44.80                 & \multicolumn{1}{c|}{47.92}                        & \multicolumn{1}{c|}{50.27}       & –                   & \multicolumn{1}{c|}{–}                   & –            \\
\multicolumn{1}{c|}{Lumina-DiMOO}             & –                 & –                    & \multicolumn{1}{c|}{–}                            & \multicolumn{1}{c|}{–}           & 51.44               & \multicolumn{1}{c|}{45.61}               & 48.54        \\
\multicolumn{1}{c|}{LightFusion-World}             & 66.69                 & 63.50                    & \multicolumn{1}{c|}{52.38}                            & \multicolumn{1}{c|}{61.85}           & –               & \multicolumn{1}{c|}{–}               & –        \\ 
\multicolumn{1}{c|}{Qwen-Image-Edit}             & –                 & –                    & \multicolumn{1}{c|}{–}                            & \multicolumn{1}{c|}{–}           &  70.95               & \multicolumn{1}{c|}{41.92}               & 56.52        \\\midrule
\rowcolor[HTML]{DCEBFA}\multicolumn{8}{c}{Unified Understanding and Generation w/ Reasoning.}                                                                                                                                                                              \\ \midrule
\multicolumn{1}{c|}{BAGEL\textsuperscript{*}}                    &66.18       &61.92          & \multicolumn{1}{c|}{49.02}                  & \multicolumn{1}{c|}{60.18} & 56.80              & \multicolumn{1}{c|}{45.10}              &50.96      \\
\multicolumn{1}{c|}{UniCoT\textsuperscript{†}}                   & \textbf{71.85}           & 67.16              & \multicolumn{1}{c|}{\textbf{63.68}}                      & \multicolumn{1}{c|}{68.00}        & –                   & \multicolumn{1}{c|}{–}                   & –            \\
\multicolumn{1}{c|}{Ours\textsuperscript{†}}                     & 70.67       &\textbf{72.38}          & \multicolumn{1}{c|}{56.89}                  & \multicolumn{1}{c|}{\textbf{68.23}} & \textbf{74.82}         & \multicolumn{1}{c|}{\textbf{65.30}}         & \textbf{70.06} \\ \midrule
\end{tabular}}
\label{table:reason_edit}
\end{table*}

\section{Experiments}
\label{Experiments}

\begin{table*}[th!]
\centering
\caption{Comparison of different models across general image generation and editing benchmarks. Bold entries represent the best performance among open-source models and underlined entries indicate the best performance among unified models with reasoning.
}
\tablestyle{1pt}{1.1}
\setlength\tabcolsep{14pt}
\resizebox{0.88\textwidth}{!}{
\begin{tabular}{lccccc}
\toprule
\multirow{2}{*}{Type}          & \multicolumn{1}{c|}{\multirow{2}{*}{Model}} & \multicolumn{2}{c|}{General T2I Generation} & \multicolumn{2}{l}{General Image Editing} \\
                               & \multicolumn{1}{c|}{}                       & GenEval       & \multicolumn{1}{c|}{DPGBench} & ImgEdit             & GEdit-EN            \\ \midrule
\multirow{3}{*}{Closed-source} & \multicolumn{1}{c|}{GPT-4o}                 & 0.84          & \multicolumn{1}{c|}{85.15}    & 4.20                & 7.53                \\
                               & \multicolumn{1}{c|}{Gemini 2.0}             & –             & \multicolumn{1}{c|}{–}        & –                   & 6.32                \\
                               & \multicolumn{1}{c|}{Seedream 4.0}           & 0.84          & \multicolumn{1}{c|}{88.25}    & 4.18                & 7.68                \\ \midrule
\rowcolor[HTML]{F6D6D3}\multicolumn{6}{c}{Unified Understanding and Generation w/o Reasoning.}                                                                                                                                        \\ \midrule
\multirow{15}{*}{Open-source}  & \multicolumn{1}{c|}{TokenFlow-XL}           & 0.55          & \multicolumn{1}{c|}{73.38}    & –                   & –                   \\
                               & \multicolumn{1}{c|}{Harmon}                 & 0.76          & \multicolumn{1}{c|}{–}        & –                   & –                   \\
                               & \multicolumn{1}{c|}{Show-o}                 & 0.53          & \multicolumn{1}{c|}{67.48}    & –                   & –                   \\
                               & \multicolumn{1}{c|}{Janus Pro}              & 0.80          & \multicolumn{1}{c|}{84.19}    & –                   & –                   \\
                               & \multicolumn{1}{c|}{MetaQuery-XL}           & 0.80          & \multicolumn{1}{c|}{82.05}    & –                   & –                   \\
                               & \multicolumn{1}{c|}{BLIP3-o}                & 0.84          & \multicolumn{1}{c|}{81.60}    & –                   & –                   \\
                               & \multicolumn{1}{c|}{UniWorld-V1}            & 0.80          & \multicolumn{1}{c|}{81.38}    & 3.26                & 4.85                \\
                               & \multicolumn{1}{c|}{Mogao}                  & 0.89          & \multicolumn{1}{c|}{84.33}    & –                   & –                   \\
                               & \multicolumn{1}{c|}{OmniGen2}               & 0.80          & \multicolumn{1}{c|}{83.57}    & 3.43                & 6.41                \\
                               & \multicolumn{1}{c|}{MMaDA}                  & 0.63          & \multicolumn{1}{c|}{69.97}    & –                   & –                   \\
                               & \multicolumn{1}{c|}{Lumina-DiMOO}           & 0.88          & \multicolumn{1}{c|}{86.04}    & –                   & –                   \\
                               & \multicolumn{1}{c|}{LightFusion-World}      & –             & \multicolumn{1}{c|}{–}        & 3.85                & 6.58                \\
                                               & \multicolumn{1}{c|}{Hunyuan-Image 3.0}      & 0.72             & \multicolumn{1}{c|}{86.10}        & –                & –                \\
                               & \multicolumn{1}{c|}{Qwen-Image}             & 0.87          & \multicolumn{1}{c|}{\textbf{88.32}}    & –                   & –                   \\
                               & \multicolumn{1}{c|}{Qwen-Image-Edit}        & –             & \multicolumn{1}{c|}{–}        & \textbf{4.27}                & \textbf{7.56}                \\ \midrule
\rowcolor[HTML]{DCEBFA}\multicolumn{6}{c}{Unified Understanding and Generation w Reasoning.}                                                                                                               \\ \midrule
\multirow{7}{*}{Open-source}   & \multicolumn{1}{c|}{T2I-R1}                 & –             & \multicolumn{1}{c|}{–}        & –                   & –                   \\
                               & \multicolumn{1}{c|}{GoT}                    & 0.64          & \multicolumn{1}{c|}{–}        & –                   & –                   \\
                               & \multicolumn{1}{c|}{Mind-Omni}              & 0.83          & \multicolumn{1}{c|}{82.50}    & –                   & –                   \\
                               & \multicolumn{1}{c|}{IRG}                    & 0.85          & \multicolumn{1}{c|}{–}        & –                   & –                   \\
                               & \multicolumn{1}{c|}{BAGEL}                  & 0.88          & \multicolumn{1}{c|}{85.07}    & 3.20                & 6.52                \\
                               & \multicolumn{1}{c|}{UniCoT}                 & 0.83          & \multicolumn{1}{c|}{–}        & –                   & 6.74                \\
                               & \multicolumn{1}{c|}{Ours}                   & \textbf{0.90} & \multicolumn{1}{c|}{\underline{86.21}}    & \underline{4.06}                & \underline{6.94}                \\ \midrule
\end{tabular}}
\label{table:general}
\end{table*}

\subsection{Experimental Setup}
\paragraph{Training Details}
In the first stage, the training corpus comprises nearly 7 million T2I samples and 500k image editing samples collected from open-source datasets including BLIP-3o~\cite{chen2025blip3ofamilyfullyopen}, ShareGPT-4o-Image~\cite{chen2025sharegpt4oimagealigningmultimodalmodels}, Echo-4o-Image~\cite{ye2025echo4oharnessingpowergpt4o}, OpenGPT4o-Image~\cite{chen2025opengpt4oimagecomprehensivedatasetadvanced}, Nano-banana-consist~\cite{nano_banana_150k}, and Pico-banana~\cite{qian2025picobanana400klargescaledatasettextguided}. We train the model's generation branch for 30,000 iterations using the Adam optimizer with a cosine learning rate schedule, including 3,000 warm-up steps, a maximum learning rate of $ 5 \times 10^{-5} $ and a minimum learning rate of $ 1 \times 10^{-5} $. 

In the second stage, the training corpus consists of 150k self-constructed single-turn knowledge-enhanced reasoning samples for T2I generation, 100k image editing reasoning samples~\cite{han2025unireditbenchunifiedreasoningbasedimage}, 
and self-constructed interleaved reasoning samples, including 36k for T2I generation and 10k for image editing.
We fine-tune all model parameters for 10,000 iterations with 1,000 warm-up steps, a maximum learning rate of $ 2 \times 10^{-5} $ and a minimum learning rate of $ 1 \times 10^{-6} $. Loss weights are set to $ \lambda_{\text{text}} = 2 $ and $ \lambda_{\text{img}} = 1 $, with a packed sequence length of 50k tokens.

\paragraph{Evaluation Setup} We evaluate world knowledge reasoning and fine-grained semantic alignment for T2I generation using the WISE~\cite{niu2025wiseworldknowledgeinformedsemantic} benchmark, which comprises 1,000 world knowledge-informed prompts across culture, natural science, and spatial and temporal comprehension. For image editing, we use UniREditBench~\cite{han2025unireditbenchunifiedreasoningbasedimage} with 2,700 meticulously curated samples covering both real- and game-world
scenarios~\cite{tong2025game0rl0}, and KrisBench~\cite{wu2025krisbenchbenchmarkingnextlevelintelligent} with 1,267 samples across factual, conceptual, and procedural knowledge to assess world knowledge reasoning and refinement capabilities. 
Additionally, we evaluate general compositional and instruction-following abilities using GenEval~\cite{ghosh2023genevalobjectfocusedframeworkevaluating} and DPGBench~\cite{hu2024ellaequipdiffusionmodels} for T2I generation, as well as  ImgEdit~\cite{ye2025imgeditunifiedimageediting} and GEdit-EN~\cite{liu2025step1xeditpracticalframeworkgeneral} for image editing.

\subsection{Main Results}
We present a comprehensive comparison of our model against existing state-of-the-art unified multimodal models that support both generation and understanding in Tab.~\ref{table:wise} and Tab.~\ref{table:reason_edit}, for world knowledge-intensive T2I generation and image editing tasks, respectively. Detailed descriptions of the compared models are provided in Appendix~\ref{sec:baseline}. 

Our model achieves the best overall performance among open-source unified multimodal models, with or without explicit reasoning mechanisms, across knowledge-intensive image generation and editing tasks. Besides, it demonstrates comparable results to closed-source models, including Seedream 4.0~\cite{ByteDanceSeedream4.0} and GPT-4o~\cite{OpenAIGPTImage1} on T2I generation, and even surpasses Gemini 2.0~\cite{kampf2025experiment} 
on KrisBench~\cite{wu2025krisbenchbenchmarkingnextlevelintelligent} and outperforms Seedream 4.0~\cite{ByteDanceSeedream4.0} on UniREditBench~\cite{han2025unireditbenchunifiedreasoningbasedimage}. These results highlight the effectiveness of our unified reasoning framework. 

Moreover, as shown in the fine-grained breakdown of performance across different knowledge domains in Tab.~\ref{table:wise} and~\ref{table:reason_edit}, our model exhibits broad and consistent world-knowledge coverage. Notably, it achieves the highest performance in \textit{Cultural Commonsense}, \textit{Spatial Reasoning}, \textit{Natural Science} including \textit{Physics} and \textit{Chemistry}. For image editing tasks, it also demonstrates strong performance across diverse knowledge categories in both KrisBench and UniREditBench. Overall, our model's knowledge-enhanced reasoning capabilities cover a wide range of tasks and domains.

\subsection{General Ability Retention}
Beyond knowledge-intensive tasks, our model remains highly competitive on general image generation and editing benchmarks while improving knowledge-enhanced reasoning, demonstrating strong generalization capability. As shown in Tab.~\ref{table:general}, on GenEval~\cite{ghosh2023genevalobjectfocusedframeworkevaluating}, our model surpasses leading systems, including Qwen-Image~\cite{wu2025qwenimagetechnicalreport}, GPT-4o~\cite{chen2025opengpt4oimagecomprehensivedatasetadvanced}, and Seedream~4.0~\cite{ByteDanceSeedream4.0}, without relying on any external LLM-based rewriting. On DPGBench~\cite{hu2024ellaequipdiffusionmodels}, it achieves the best performance among models with reasoning mechanisms during generation, highlighting strong long-horizon instruction following. We further evaluate precise instruction-following image editing on ImgEdit~\cite{ye2025imgeditunifiedimageediting} and GEdit-EN~\cite{liu2025step1xeditpracticalframeworkgeneral}, which are essential for practical refinement. Our model delivers the strongest results among models with reasoning capability while remaining competitive with a broad range of existing approaches. These results indicate that our model is not only strong in reasoning-centric settings but also excels in general generation and editing, providing a robust and versatile unified foundation. Detailed results are shown in Appendix~\ref{sec:detaild results} and case studies are shown in Appendix~\ref{sec:case}.

\subsection{Ablation Study}
\begin{table}[t]
\centering
\caption{Ablation study of \textbf{UniReason}. The base model is BAGEL~\cite{deng2025emergingpropertiesunifiedmultimodal}. ``Two-Stage Training'' refers to fine-tuning the base model using the two-stage training recipe, as described in Sec.~\ref{sec:training recipe}.}
\tablestyle{0pt}{1.3}
\setlength\tabcolsep{5pt}{
\resizebox{0.49\textwidth}{!}{
\begin{tabular}{l|c|c|c}
\hline
Method                                      & WISE & KrisBench & UniREditBench \\ \midrule
Base Model                                  & 0.52 & 56.21     & 50.96         \\ \midrule
+ Two-Stage Training                          & 0.58\gain{(+0.06)} & 61.53\gain{(+5.32)}     & 63.37\gain{(+12.41)}         \\
+ Reasoning  & 0.73\gain{(+0.21)} & 64.12\gain{(+7.91)}     & 67.30\gain{(+16.34)}          \\
+ Refinement & \textbf{0.78 \gain{(+0.26)}} & \textbf{68.23\gain{(+12.02)}}    & \textbf{70.06\gain{(+19.10)}}          \\ \midrule
\end{tabular}
}
}
\label{table:Ablation}
\end{table}
We further investigate the contributions of the two-stage training strategy, as well as the reasoning and refinement mechanisms for image synthesis. On three knowledge-intensive generation and editing benchmarks, we compare three progressive settings built upon the BAGEL base model: (i) \emph{Two-Stage Training}, which performs direct image generation after two-stage fine-tuning; (ii) \emph{+ Reasoning}, which elicits textual reasoning prior to image synthesis; and (iii) \emph{+ Refinement}, which further introduces an explicit reflection and refinement step to produce a final refined output. 

Tab.~\ref{table:Ablation} shows consistent improvement across all benchmarks as each component is added. The two-stage training alone effectively improves the base model's instruction-following and synthesis capabilities. Then, introducing world knowledge-enhanced textual reasoning yields significant gains, especially on WISE with a +0.21 improvement. Finally, the visual refinement phase further improves the overall performance on all benchmarks. These results suggest that the two-stage training strategy injects both knowledge-enhanced reasoning and fine-grained refinement capabilities into the unified multimodal model, rather than merely enhancing surface-level visual composition. Moreover, the results highlight the importance of explicitly modeling implicit world knowledge during initial synthesis and performing fine-grained editing for further refinement.

\subsection{Correlation of Editing and Refinement}
\begin{figure}[t]
    \centering
    \includegraphics[width=1.05\linewidth]{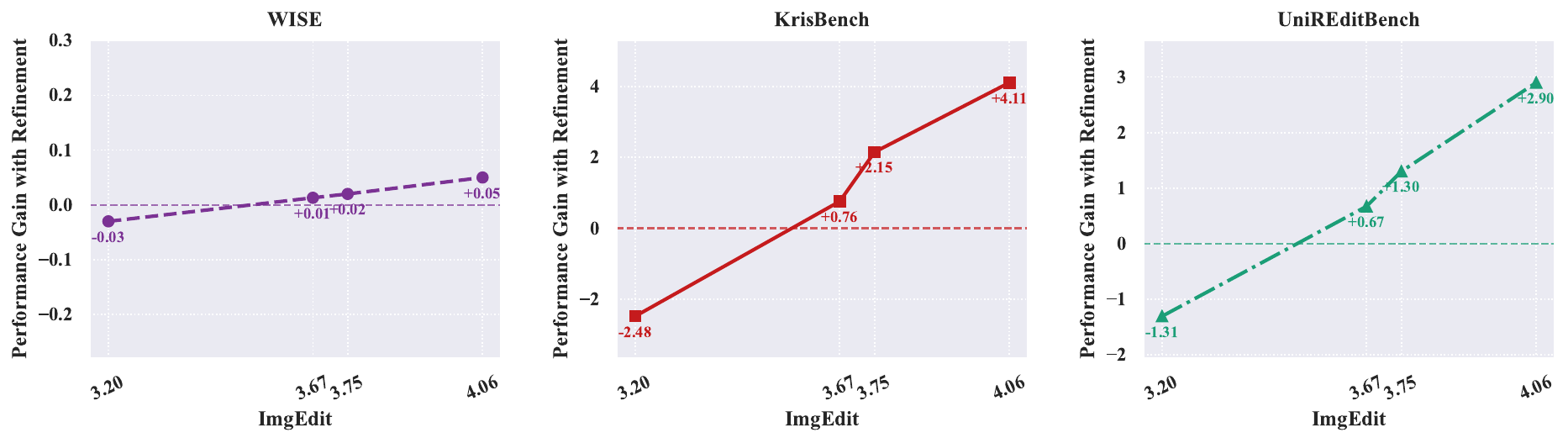} 
    \caption{Correlation between image editing capability (ImgEdit score) and performance gains from refinement across three benchmarks. Higher editing proficiency leads to monotonically increasing refinement effectiveness.}
    \label{fig:refine}
\end{figure}

To show how image editing capability affects refinement effectiveness, we analyze performance gains with and without the refinement mechanism across models with varying editing capabilities. Specifically, we select different checkpoints during stage-1 training, each exhibiting different levels of editing proficiency, and apply identical stage-2 training to all checkpoints. We then evaluate performance on three knowledge-intensive benchmarks, measuring the gains achieved through refinement after initial textual reasoning. Fig.~\ref{fig:refine} plots these performance gains against the editing performance of each checkpoint on ImgEdit.

The results reveal that performance gains from refinement increase monotonically with higher ImgEdit scores. This trend highlights the importance of jointly training image editing and T2I generation within a unified interleaved reasoning framework that integrates both textual reasoning and visual refinement. Since visual refinement relies on fine-grained and controllable editing, insufficient editing capacity can limit the effectiveness of reasoning-guided refinement.

\section{Conclusion}

In this paper, we introduce \textbf{\modelname}, a unified reasoning framework that harmonizes the text-to-image generation and image editing by exploiting their inherent structural synergies. Specifically, we proposed two complementary components: World Knowledge-Enhanced Textual Reasoning that infers implicit common sense and physical laws, and Fine-grained Editing-like Visual Refinement that enables iterative reflection and correction. By constructing high-quality datasets across five knowledge categories and employing a two-stage training strategy, \textbf{\modelname} demonstrates superior instruction following and visual fidelity. Extensive experiments on multiple benchmarks demonstrate that our unified reasoning approach achieves advanced performance across both T2I and editing tasks.

\section{Impact Statement}
This work focuses on improving reasoning and alignment in image generation and editing models. While such advances may benefit various creative and assistive applications, they may also introduce risks related to misuse of generated visual content. Addressing these risks requires system-level safeguards and responsible deployment practices beyond the scope of this paper.

\bibliographystyle{unsrtnat}  
\bibliography{unireason} 

\clearpage

\beginappendix
\section{Appendix}

\subsection{Compared Baselines}
\label{sec:baseline}
We compared closed-source models including: GPT-4o~\cite{OpenAIGPTImage1}, Gemini-2.0~\cite{google_gemini25_pro_2025}, Seedream4.0~\cite{ByteDanceSeedream4.0}, as well as open-source advanced unified
multimodal models models which support both multimodal understanding and high quality image generation including autoregressive unified models, such as Harmon~\cite{wu2025harmonizing}, TokenFlow-XL~\cite{qu2025tokenflowunifiedimagetokenizer}, and Janus-Pro~\cite{chen2025janusprounifiedmultimodalunderstanding}. Discrete diffusion-based approaches, including Lumina-DiMOO~\cite{xin2025luminadimooomnidiffusionlarge}, MMaDA~\cite{yang2025mmadamultimodallargediffusion}, and Show-o~\cite{xie2025showosingletransformerunify}. Another line of work connects VLMs and diffusion transformers via explicit connectors, exemplified by BLIP-3o~\cite{chen2025blip3ofamilyfullyopen}, UniWorld-V1~\cite{lin2025uniworldv1highresolutionsemanticencoders}, OmniGen2~\cite{wu2025omnigen2explorationadvancedmultimodal}, and the Qwen-Image series~\cite{wu2025qwenimagetechnicalreport}. In contrast, deep fusion methods tightly integrate VLMs and DiTs within a unified architecture, such as Mogao~\cite{liao2025mogaoomnifoundationmodel}, Hunyuan Image~3.0~\cite{cao2025hunyuanimage30technicalreport} and LightFusion-World~\cite{gou2025vqvaworldhighqualityvisual}, the latter further enhanced with knowledge-centric fine-tuning.

Among open-source unified multimodal models that support naive reasoning, T2I-R1~\cite{jiang2025t2ir1reinforcingimagegeneration}, MindOmni~\cite{xiao2025mindomniunleashingreasoninggeneration}, and BAGEL~\cite{deng2025emergingpropertiesunifiedmultimodal} primarily rely on textual reasoning to decompose abstract instructions into explicit semantic components that guide image generation. In contrast, GoT~\cite{fang2025gotunleashingreasoningcapability} introduces coordinate-based representations to provide explicit spatial guidance during synthesis. Another line of work, including IRG~\cite{huang2025interleavingreasoningbettertexttoimage} and UniCoT~\cite{qin2025unicotunifiedchainofthoughtreasoning}, adopts interleaved reasoning mechanisms to reorganize semantics across modalities, progressively decomposing instructions into finer-grained and more structured descriptions for generation and refinement.

\subsection{Data Preparation Details}
\label{sec:data_details}
To construct high-quality supervision for training \textbf{UniReason} across both text-to-image (T2I) generation and image editing tasks, we design a two-phase data construction pipeline that integrates \emph{world knowledge–enhanced textual reasoning} with \emph{fine-grained editing-like visual refinement}. 

\paragraph{Phase I: World Knowledge–Enhanced Reasoning Data Construction} We first build challenging instructions that require reasoning beyond pixel-level completion, covering five categories of world knowledge: (i) Cultural Commonsense, which resolves culturally grounded but underspecified entities using shared knowledge of history, customs, and symbols; (ii) Natural Science, which enforces consistency with physical, biological, medical, or chemical laws; (iii) {Spatial Reasoning}, which derives correct relative positions, orientations, viewpoints, and camera transformations; (iv) Temporal Reasoning, which models time-dependent state transitions and causal event sequences; and (v) Logical Reasoning, which translates explicit or implicit logical constraints into visually valid solutions. For T2I generation, we manually curate seed prompts grounded in Wikipedia and category definitions, then use Gemini-2.5 Pro~\cite{google_gemini25_pro_2025} to expand them and generate corresponding textual CoT reasoning. Each prompt–reasoning pair is rendered into images using Qwen-Image~\cite{wu2025qwenimagetechnicalreport}, forming reasoning-grounded training samples. For image editing, we adopt triplets from UniREdit-Data-100K~\cite{han2025unireditbenchunifiedreasoningbasedimage}, augmented with Gemini-2.5 Pro–generated reasoning processes with category definitions. All samples are filtered by Gemini-2.5 Pro to ensure instruction alignment, visual fidelity, and knowledge-consistent reasoning, retaining only verified high-quality data. 

\paragraph{Phase II: Fine-grained Editing-like Visual Refinement Data Construction} To further train interleaved reasoning and refinement capabilities, we design an agent-based pipeline to generate iterative refinement supervision. Given an input instruction, an initial generator produces a draft image along with textual reasoning. A verifier (Gemini-2.5 Pro) then diagnoses caption–image mismatches and outputs structured, actionable feedback along five dimensions: object presence, attribute accuracy, style consistency, realism, and aesthetic quality. A refinement teacher (Qwen-Image-Edit~\cite{wu2025qwenimagetechnicalreport}) applies this feedback and textual reasoning via instruction-guided image editing to produce a refined image. Finally, a judge (Gemini-2.5 Pro) performs comparative evaluation between the initial and refined images, retaining refined results only if they demonstrate measurable improvements and faithfully reflect the suggested modifications. Concretely, we sample long-form captions from ShareGPT-4o-Image~\cite{chen2025sharegpt4oimagealigningmultimodalmodels} and short-form captions from Midjourney prompts\footnote{https://huggingface.co/datasets/vivym/midjourney-prompts} for T2I generation, and image–instruction pairs from UniREdit-Data-100K for image editing. These inputs undergo the full generation–verification–refinement–selection cycle, yielding a high-quality training set that jointly supports world knowledge–enhanced reasoning and fine-grained visual refinement.

\subsection{Detailed Evaluation Results}
\label{sec:detaild results}
We show the detailed evaluation results on general tasks include GenEval~\cite{ghosh2023genevalobjectfocusedframeworkevaluating} shown in Tab.~\ref{table:geneval} and DPGBench~\cite{hu2024ellaequipdiffusionmodels} in Tab.~\ref{table:dpg} for T2I generation, as well as  ImgEdit~\cite{ye2025imgeditunifiedimageediting} and GEdit-EN~\cite{liu2025step1xeditpracticalframeworkgeneral} in Tab.~\ref{table:edit} for image editing. The results show our model delivers the strongest results among
models with reasoning while remaining competitive with
a broad range of existing approaches. These results indicate that our model is not only strong in reasoning-centric
settings but also excels in general generation and editing,
providing a robust and versatile unified foundation.

\begin{table*}[th]
\centering
\caption{Evaluation of general text-to-image generation capabilities on GenEval~\cite{ghosh2023genevalobjectfocusedframeworkevaluating} benchmark.}
\tablestyle{8pt}{1.0}
\setlength\tabcolsep{8pt}
\resizebox{1\textwidth}{!}{
\begin{tabular}{cccccccc}
\toprule
\multicolumn{1}{c|}{Model}       & Single object & Two object & Counting & Colors   & Position & \multicolumn{1}{c|}{Attribution} & Overall  \\ \midrule
\multicolumn{1}{c|}{GPT-4o}       & 0.99           & 0.92        & 0.85      & 0.92      & 0.75      & \multicolumn{1}{c|}{0.61}         & 0.84      \\ 
\multicolumn{1}{c|}{Seedream 4.0}       & 0.99           & 0.92        &  0.72      & 0.91      & 0.76      & \multicolumn{1}{c|}{0.74}         & 0.84      \\ \midrule
\rowcolor[HTML]{F6D6D3}\multicolumn{8}{c}{Unified Understanding and Generation w/o Reasoning.}                                                                                  \\ \midrule
\multicolumn{1}{c|}{TokenFlow-XL} & 0.95           & 0.60         & 0.41      & 0.81      & 0.16      & \multicolumn{1}{c|}{0.24}         & 0.55      \\
\multicolumn{1}{c|}{Harmon}       & 0.99           & 0.86        & 0.66      & 0.85      & 0.74      & \multicolumn{1}{c|}{0.48}         & 0.76      \\
\multicolumn{1}{c|}{Show-o}      & 0.95              & 0.52       & 0.49      & 0.82      & 0.11      & \multicolumn{1}{c|}{0.28}         & 0.53      \\
\multicolumn{1}{c|}{Janus Pro}    & 0.99           & 0.89        & 0.59      & 0.90       & 0.79      & \multicolumn{1}{c|}{0.66}         & 0.80       \\

\multicolumn{1}{c|}{MetaQuery-XL} & –              & –           & –         & –         & –         & \multicolumn{1}{c|}{–}            & 0.80       \\
\multicolumn{1}{c|}{BLIP3-o}      & –              & –           & –         & –         & –         & \multicolumn{1}{c|}{–}            & 0.84      \\
\multicolumn{1}{c|}{UniWorld-V1}  & 0.99           & 0.93        & 0.79      & 0.89      & 0.49      & \multicolumn{1}{c|}{0.70}          & 0.80       \\
\multicolumn{1}{c|}{Mogao}        & 1.00              & 0.97        & 0.83      & 0.93      & 0.84      & \multicolumn{1}{c|}{0.80}          & 0.89      \\
\multicolumn{1}{c|}{OmniGen2}     & 1.00              & 0.95        & 0.64      & 0.88      & 0.55      & \multicolumn{1}{c|}{0.76}         & 0.80       \\
\multicolumn{1}{c|}{MMaDA}        & 0.99           & 0.76        & 0.61      & 0.84      & 0.20       & \multicolumn{1}{c|}{0.37}         & 0.63      \\
\multicolumn{1}{c|}{Lumina-DiMOO} & 1.00              & 0.94        & 0.85      & 0.89      & 0.85      & \multicolumn{1}{c|}{0.76}         & 0.88      \\ 
\multicolumn{1}{c|}{Hunyuan-Image 3.0} & 1.00              & 0.92        & 0.48      & 0.82      & 0.42      & \multicolumn{1}{c|}{0.63}         & 0.72     \\
\multicolumn{1}{c|}{Qwen-Image} & 0.99              & 0.92        & 0.89      &  0.88      & 0.76      & \multicolumn{1}{c|}{0.77}         & 0.87      \\ \midrule
\rowcolor[HTML]{DCEBFA}\multicolumn{8}{c}{Unified Understanding and Generation w Reasoning.}                                                                                  \\ \midrule
\multicolumn{1}{c|}{GoT}          & –           & –        & –      & –      & –      & \multicolumn{1}{c|}{–}         & 0.64      \\
\multicolumn{1}{c|}{Mind-Omni}          & 0.99          & 0.94        & 0.71      & 0.90     & 0.71      & \multicolumn{1}{c|}{0.71}         & 0.83      \\
\multicolumn{1}{c|}{IRG}          & 0.98           & 0.94      & 0.83      & 0.86      & 0.74      & \multicolumn{1}{c|}{0.73}         & 0.85     \\
\multicolumn{1}{c|}{BAGEL}        & 0.98      & 0.95   & 0.84 & 0.95 & 0.78 & \multicolumn{1}{c|}{0.77}   & 0.88 \\
\multicolumn{1}{c|}{Uni-CoT}   & 0.99      & 0.96   & 0.84 & 0.92  & 0.57 & \multicolumn{1}{c|}{0.71}    & 0.83 \\
\multicolumn{1}{c|}{Ours}         & 1.00         & 0.96   & 0.82 & 0.90 & 0.88 & \multicolumn{1}{c|}{0.82}    & 0.90 \\ \midrule

\end{tabular}}
\label{table:geneval}
\end{table*}

\begin{table*}[th!]
\centering
\caption{Evaluation of general text-to-image generation capabilities on DPG~\cite{ghosh2023genevalobjectfocusedframeworkevaluating} benchmark.}
\tablestyle{12pt}{1.0}
\setlength\tabcolsep{12pt}
\resizebox{0.9\textwidth}{!}{
\begin{tabular}{lcccccc}
\toprule
\multicolumn{1}{l|}{Model}       & Global & Entity & Attribute & Relation & \multicolumn{1}{c|}{Other} & Overall       \\ \midrule
\multicolumn{1}{l|}{GPT-4o}       & 88.89   & 88.94   & 89.84      & 92.63     & \multicolumn{1}{c|}{90.96}  & 85.15          \\ 
\multicolumn{1}{l|}{Seedream 4.0}       & 94.10   & 92.28   &  92.75      & 93.67     & \multicolumn{1}{c|}{92.77}  & 88.25          \\\midrule
\rowcolor[HTML]{F6D6D3}\multicolumn{7}{c}{Unified Understanding and Generation w/o Reasoning.}                                                                     \\ \midrule
\multicolumn{1}{l|}{TokenFlow-XL} & 78.72   & 79.22   & 81.29      & 85.22     & \multicolumn{1}{c|}{71.20}   & 73.38          \\
\multicolumn{1}{l|}{Show-o}      & –      & –   & –      & –     & \multicolumn{1}{c|}{–}  & 67.48          \\
\multicolumn{1}{l|}{Janus Pro}    & 86.90    & 88.90    & 89.40       & 89.32     & \multicolumn{1}{c|}{89.02}  & 84.19          \\
\multicolumn{1}{l|}{MetaQuery-XL} & –       & –       & –          & –         & \multicolumn{1}{c|}{–}      & 82.05          \\
\multicolumn{1}{l|}{BLIP3-o}      & –       & –       & –          & –         & \multicolumn{1}{c|}{–}      & 81.60           \\
\multicolumn{1}{l|}{UniWorld-V1}  & 83.64   & 88.39   & 88.44      & 89.27     & \multicolumn{1}{c|}{87.22}  & 81.38          \\
\multicolumn{1}{l|}{Mogao}        & 82.37   & 90.03   & 88.26      & 93.18     & \multicolumn{1}{c|}{85.40}   & 84.33          \\
\multicolumn{1}{l|}{OmniGen2}     & 88.81   & 88.83   & 90.18      & 89.37     & \multicolumn{1}{c|}{90.27}  & 83.57          \\
\multicolumn{1}{l|}{MMaDA}        & 77.81   & 78.48   & 81.74      & 84.79     & \multicolumn{1}{c|}{63.20}   & 69.97          \\
\multicolumn{1}{l|}{Lumina-DiMOO} & 81.46   & 92.08   & 88.98      & 94.31     & \multicolumn{1}{c|}{82.00}     & 86.04          \\
\multicolumn{1}{l|}{Hunyuan-Image 3.0}  & 92.12   & 92.53   &  89.13      &  92.13     & \multicolumn{1}{c|}{91.92}  & 86.10          \\ 
\multicolumn{1}{l|}{Qwen-Image}  & 91.32   & 91.56   & 92.02      & 94.31     & \multicolumn{1}{c|}{92.73}  & 88.32          \\ \midrule
\rowcolor[HTML]{DCEBFA}\multicolumn{7}{c}{Unified Understanding and Generation w Reasoning.}                                            \\ \midrule
\multicolumn{1}{l|}{Mind-Omni}        & 89.10   & –   & –      & –     & \multicolumn{1}{c|}{89.20}  & 82.50          \\
\multicolumn{1}{l|}{BAGEL}        & 88.94   & 90.37   & 91.29    & 90.82     & \multicolumn{1}{c|}{88.67}  & 85.07          \\
\multicolumn{1}{l|}{Ours}         & 
91.78   & 91.23   & 90.76   & 91.12     & \multicolumn{1}{c|}{92.27}  & 86.21 \\ \midrule
\end{tabular}}
\label{table:dpg}
\end{table*}

\begin{table*}[th!]
\centering
\caption{Evaluation of general image editing capabilities on ImgEdit~\cite{ye2025imgeditunifiedimageediting} and GEdit-EN~\cite{liu2025step1xeditpracticalframeworkgeneral} benchmarks.}
\tablestyle{4pt}{1.6}
\setlength\tabcolsep{4pt}
\resizebox{1\textwidth}{!}{
\begin{tabular}{cccccccccccccc}
\toprule
\multicolumn{1}{c|}{}                                  & \multicolumn{10}{c|}{ImgEdit}                                                                                                                                                                                                                                                        & \multicolumn{3}{c}{GEdit-EN} \\ \cline{2-14} 
\multicolumn{1}{c|}{\multirow{-2}{*}{Model}}          & Add                         & Adjust                      & Extract                    & Replace                     & Remove                      & Background                  & Style                       & Hybrid & \multicolumn{1}{c|}{Action} & \multicolumn{1}{c|}{Overall} & G\_SC    & G\_PQ    & G\_O   \\ \cline{1-14}
\multicolumn{1}{c|}{{\color[HTML]{1F2329} GPT-4o}}     & {\color[HTML]{1F2329} 4.61} & {\color[HTML]{1F2329} 4.33} & {\color[HTML]{1F2329} 2.90} & {\color[HTML]{1F2329} 4.35} & {\color[HTML]{1F2329} 3.66} & {\color[HTML]{1F2329} 4.57} & {\color[HTML]{1F2329} 4.93} & 3.96   & \multicolumn{1}{c|}{4.89}   & \multicolumn{1}{c|}{4.20}     & 7.85     & 7.62     & 7.53   \\
\multicolumn{1}{c|}{{\color[HTML]{1F2329} Gemini 2.0}} & {\color[HTML]{1F2329} –}    & {\color[HTML]{1F2329} –}    & {\color[HTML]{1F2329} –}   & {\color[HTML]{1F2329} –}    & {\color[HTML]{1F2329} –}    & {\color[HTML]{1F2329} –}    & {\color[HTML]{1F2329} –}    & –      & \multicolumn{1}{c|}{–}      & \multicolumn{1}{c|}{–}       & 6.73     & 6.61     & 6.32   \\ 
\multicolumn{1}{c|}{{\color[HTML]{1F2329} Seedream 4.0}} & {\color[HTML]{1F2329} 4.52}    & {\color[HTML]{1F2329} 4.41}    & {\color[HTML]{1F2329} 2.93}   & {\color[HTML]{1F2329}  4.56}    & {\color[HTML]{1F2329} 4.44}    & {\color[HTML]{1F2329}  4.30}    & {\color[HTML]{1F2329} 4.76}    & 3.33      & \multicolumn{1}{c|}{ 4.36}      & \multicolumn{1}{c|}{ 4.18}       & 8.24     & 8.08     & 7.68   \\\midrule
\rowcolor[HTML]{F6D6D3}\multicolumn{14}{c}{Unified Understanding and Generation w/o Reasoning.}                                                                                                                                                                                                                                                                                                                   \\ \midrule
\multicolumn{1}{c|}{Janus 4o}                          & 3.60                         & –                           & 2.28                       & 3.27                        & 2.28                        & 3.32                        & 4.47                        & 2.74   & \multicolumn{1}{c|}{4.13}   & \multicolumn{1}{c|}{3.26}    & –        & –        & –      \\
\multicolumn{1}{c|}{UniWorld-V1}                       & 3.82                        & 3.66                        & 2.31                       & 3.45                        & 3.02                        & 2.99                        & 4.71                        & 2.96   & \multicolumn{1}{c|}{2.74}   & \multicolumn{1}{c|}{3.26}    & 4.93     & 7.43     & 4.85   \\
\multicolumn{1}{c|}{OmniGen2}                          & 3.74                        & 3.54                        & 1.77                       & 3.21                        & 2.77                        & 3.57                        & 4.81                        & 2.30    & \multicolumn{1}{c|}{4.14}   & \multicolumn{1}{c|}{3.43}    & 7.16     & 6.77     & 6.41   \\
\multicolumn{1}{c|}{LightFusion-World}                 & 4.33                           & 3.37                           & 1.25                          & 4.63                           & 3.74                           & 4.24                           & 4.69                           & 3.91      & \multicolumn{1}{c|}{4.45}      & \multicolumn{1}{c|}{3.85}    & 7.00        & 7.29     & 6.58   \\ 
\multicolumn{1}{c|}{Qwen-Image-Edit}                 & 4.38                           & 4.16                           &3.43                          & 4.66                           & 4.14                           & 4.38                           & 4.81                           & 3.82      & \multicolumn{1}{c|}{4.69}      & \multicolumn{1}{c|}{4.27}    &  8.00        & 7.86     & 7.56   \\\midrule
\rowcolor[HTML]{DCEBFA}\multicolumn{14}{c}{Unified Understanding and Generation w Reasoning.}                                                                                                                                                                                                                                                                                          \\ \midrule
\multicolumn{1}{c|}{BAGEL}                             & 3.56                        & 3.31                        & 1.88                       & 2.62                        & 2.88                        & 3.44                        & 4.49                        & 2.38   & \multicolumn{1}{c|}{4.17}   & \multicolumn{1}{c|}{3.20}     & 7.36     & 6.83     & 6.52   \\
\multicolumn{1}{c|}{UniCoT}                             & –                        & –                        & –                       & –                        & –                       & –                        & –                        & –   & \multicolumn{1}{c|}{–}   & \multicolumn{1}{c|}{–}     & 7.91     & 6.24     & 6.74   \\
\multicolumn{1}{c|}{Ours}                              & 4.14                        & 4.06               & 2.49                       & 4.42                        & 4.31                        & 4.23                        & 4.65                        & 2.58   & \multicolumn{1}{c|}{4.68}   & \multicolumn{1}{c|}{4.06}    & 7.46      & 7.66      & 6.94   \\ \midrule
\end{tabular}}
\label{table:edit}
\end{table*}
\subsection{Case Study}
\label{sec:case}
\begin{figure*}[th]
    \centering
    \includegraphics[width=0.92\linewidth]{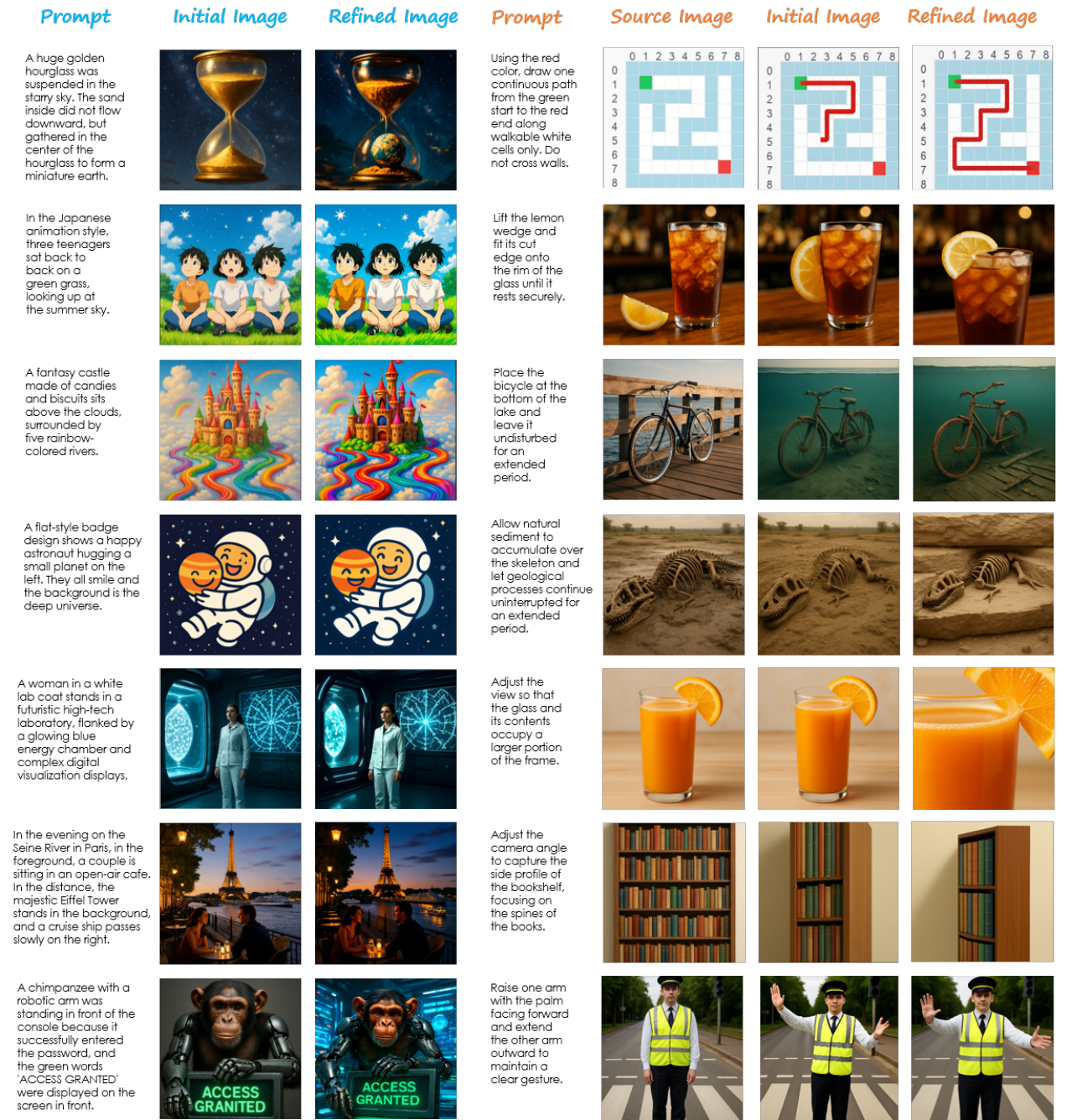} 
    \caption{Qualitative results of \textbf{UniReason} on both T2I generation (blue column) and image editing task (orange column).}
    \label{Fig:more_case}
\end{figure*}

We present additional \textbf{UniReason} results on both T2I generation and image editing tasks in Fig.~\ref{Fig:more_case}. The results demonstrate that, while maintaining high-quality T2I generation and image editing performance, \textbf{UniReason} exhibits strong reasoning capabilities, enabling it to handle complex scenarios such as maze navigation, temporal evolution, and spatial camera viewpoint transformations. Moreover, \textbf{UniReason} shows robust refinement ability, effectively correcting fine-grained details such as faces, text, and hand gestures, thereby improving the quality of the initial images and rectifying errors introduced during the initial generation.

\end{document}